\documentclass[11pt,a4paper]{article}
\usepackage[hyperref]{emnlp2018}
\usepackage{times}
\usepackage{latexsym}
\usepackage[utf8]{inputenc}
\usepackage{multirow}
\usepackage{color}
\usepackage{xcolor}
\usepackage{url}
\usepackage{graphicx}
\usepackage{caption}
\usepackage{xcolor,colortbl}
\usepackage{booktabs}
\aclfinalcopy 
\newcommand{\zied}[1]{\textcolor{black}{#1}}

\title{Analyzing Learned Representations of a Deep ASR Performance Prediction Model}

\author{ 
Zied Elloumi$^{1,2}$ \\\And Laurent Besacier$^{2}$ \\\And Olivier Galibert$^{1}$ \\  \\
\hspace{-4cm}  $^{1}$Laboratoire national de métrologie et d\textquoteright essais (LNE) , France \\
 \hspace{-4cm} $^{2}$Univ. Grenoble Alpes, CNRS, Grenoble INP, LIG, F-38000 Grenoble, France \\ 
\hspace{-4cm}   {\tt firstname.name@lne.fr} \\
\hspace{-4cm}   {\tt firstname.name@univ-grenoble-alpes.fr} \\
  \\\And
  Benjamin Lecouteux$^{2}$ \\
  }  

\date{}

\begin{document}
\maketitle
\begin{abstract}

 This paper addresses a relatively new task: prediction of  ASR performance on unseen broadcast programs. 
In a previous paper, we presented an ASR performance prediction system using CNNs that encode both text (ASR transcript) and speech, in order to predict word error rate. 
This work is  dedicated to the analysis of speech signal embeddings and text embeddings learnt by the CNN while training  our prediction model.
We try to better understand which information is captured by the deep model and its relation with different conditioning factors.
It is shown that hidden layers convey a clear signal about speech style, accent and broadcast type. We then try to leverage these 3 types of information at training time through multi-task learning. Our experiments show that this allows to train slightly more efficient ASR performance prediction systems that - in addition - simultaneously tag the analyzed utterances according to their speech style, accent and broadcast program origin.

\end{abstract}

\section{Introduction}

Predicting automatic speech recognition (ASR) performance on unseen speech recordings is an important Grail of speech research.
In a previous paper \cite{elloumi2018asr}, we presented a framework for modeling and evaluating ASR performance prediction on unseen broadcast programs. CNNs were very efficient encoding both text (ASR transcript) and speech to predict ASR word error rate (WER). 
However, while achieving state-of-the-art performance prediction results, our CNN approach is more difficult to understand compared to  conventional approaches based on  engineered features such as \textit{TransRater}\footnote{\url{https://github.com/hlt-mt/TranscRater}} for instance.  This lack of interpretability of the representations learned by deep neural networks is a general problem in AI. Recent papers started to address this issue and analyzed hidden representations learned during training of different natural language processing models \cite{mohamed2012understanding,DBLP:journals/corr/WuK16,belinkov2017analyzing,shi2016does,belinkov2017evaluating,wang2017does}.

\textbf{Contribution.} This work is  dedicated to the analysis of speech signal embeddings and text embeddings learnt by the CNN during training of our ASR performance prediction model. Our goal is to better understand which information is captured by the deep model and its relation with conditioning factors such as speech style, accent or broadcast program type. For this, we use a data set presented in \cite{elloumi2018asr} which contains a large amount of speech utterances taken from various collections of French broadcast programs. Following a methodology similar to  \cite{belinkov2017analyzing}, our deep performance prediction model is used to generate utterance level features that are given to a shallow classifier trained to solve secondary classification tasks. It is shown that hidden layers convey a clear signal about speech style, accent and show. We then try to leverage these 3 types of information at training time through multi-task learning. Our experiments show that this allows to train slightly more efficient ASR performance prediction systems that - in addition - simultaneously tag the analyzed utterances according to their speech style, accent and broadcast program origin.
 
\textbf{Outline.} The paper is organized as follows. In section~\ref{sec:RelatedWorks}, we present a brief overview of related works and present our ASR performance prediction system in section \ref{sec:icasspsystem}. Then, we detail our methodology to evaluate learned representations in section~\ref{sec:EvaluatingLR}. Our multi-task learning experiments for ASR performance prediction are presented in section~\ref{sec:multi-task}. Finally, section~\ref{sec:conclusion} concludes this work.

 \vspace{-.3cm}
\section{\label{sec:RelatedWorks}Related works}

 Several works tried to understand learned representations  for  NLP tasks such as Automatic Speech Recognition (ASR) and Neural Machine Translation (NMT).
 
\cite{shi2016does} and \cite{belinkov2017evaluating}  tried to better understand the hidden representations of NMT models which were given to a shallow classifier in order to predict syntactic labels \cite{shi2016does}, part-of-speech labels or semantic ones \cite{belinkov2017evaluating}. It was shown that lower layers are better at POS tagging, while higher layers are better at learning semantics.
\cite{mohamed2012understanding} and \cite{belinkov2017analyzing}  analyzed the feature representations from a deep ASR model using t-SNE visualization~\cite{maaten2008visualizing} and tried to understand which layers better capture the phonemic information by training a shallow phone classifier. 
Also relevant is the work of \cite{wang2017does} who proposed an in-depth investigation on three kinds of speaker embeddings learned for a speaker recognition task, i.e. i-vector, d-vector and RNN/LSTM based sequence-vector (s-vector). Classification tasks were designed to facilitate better understanding of the encoded speaker representations.  Multi-task learning was also proposed to integrate different speaker embeddings and improve speaker verification performance.

 
\section{\label{sec:icasspsystem} ASR performance prediction system}

In \cite{elloumi2018asr}, we proposed a new approach  using convolution neural networks (CNNs) to predict ASR performance from a collection of heterogeneous broadcast programs (both radio and TV). We particularly focused on the combination of text (ASR transcription) and signal (raw speech) inputs which both proved useful for CNN prediction.
We also observed that our system  remarkably predicts  WER distribution on a collection of speech recordings.
 
 To obtain speech transcripts (ASR outputs) for the prediction model, we built our own French ASR system based on the KALDI toolkit \cite{povey2011kaldi}.  
 A hybrid HMM-DNN system was trained using 100 hours of broadcast news from \textit{Quaero}\footnote{http://www.quaero.org},  \textit{ETAPE} \cite{gravier2012etape}, \textit{ESTER 1 \& ESTER 2} \cite{galliano2005ester} and  \textit{REPERE}  \cite{kahn2012presentation} collections. ASR performance was evaluated on the held out corpora presented in table  \ref{tab:DistShow} (used to train and evaluate ASR prediction) and its averaged value was 22.29\% on the TRAIN set, 22.35\% on the DEV set and 31.20\%  on the TEST set (which contains more challenging broadcast programs).

\begin{figure*}[htb]
\begin{minipage}[b]{\linewidth}
  \centering
  \centerline{\includegraphics[width=
  \linewidth]{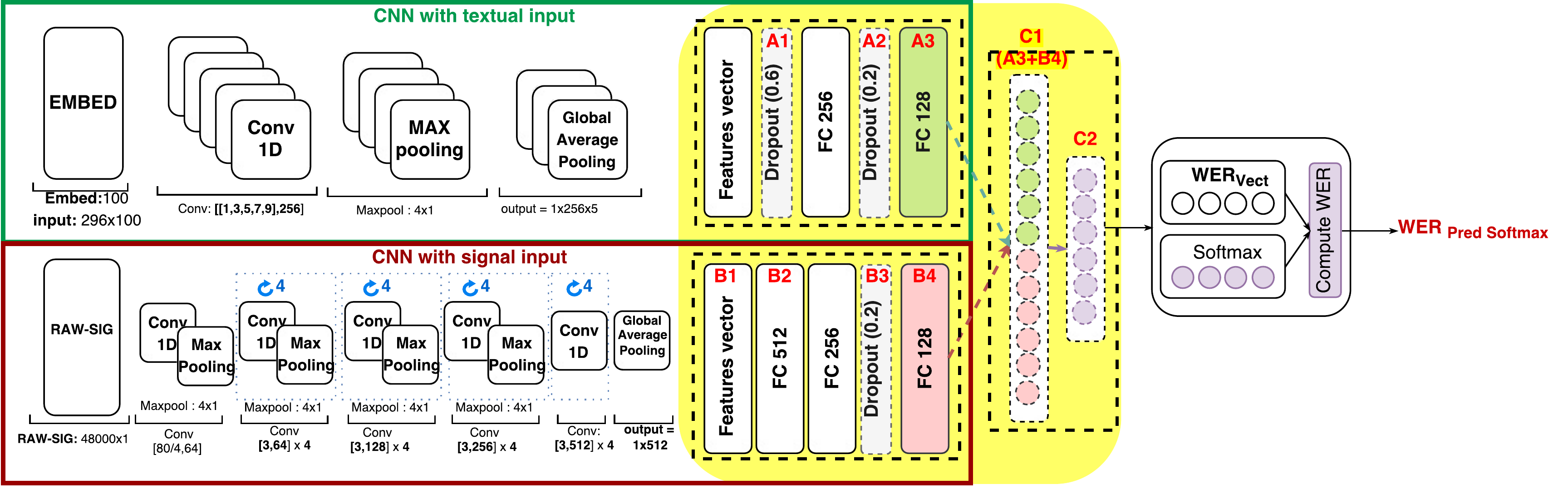}}
\end{minipage}
\caption{Architecture of our CNN with text (green) and signal (red) inputs for WER prediction}
\label{fig:architecture}
\end{figure*}

Figure~\ref{fig:architecture} shows our network architecture.
The network input can be either a pure text input, a pure signal input (raw signal) or a dual (text+speech) input. To avoid memory issues, signals are downsampled to  8khz  and models are trained on six-second speech turns (shorter speech turns are padded with zeros). For text input, the architecture is inspired from \cite{kim2014convolutional} (green in Figure~\ref{fig:architecture}): the input is a matrix of dimensions 296x100 (296 is the longest ASR hypothesis length in our corpus ; 100 is the dimension of pre-trained word embeddings on a large held out text corpus of 3.3G words). For speech input, we use the best architecture  ({\em m18}) proposed in \cite{CNNrawWav} (colored in red in Figure~\ref{fig:architecture}) of dimensions 48000 x 1 (48000 samples correspond to 6s of speech).  
 
 For WER prediction, our best approach (called CNN$_{Softmax}$) used $softmax$ probabilities and an external fixed WER$_{Vector}$ which  corresponds to a discretization of the WER output space (see \cite{elloumi2018asr} for more details). The best performance  obtained is 19.24\% MAE\footnote{Mean Absolute Error (MAE) is a common metric to evaluate WER prediction ; it computes the absolute deviation between the true and predicted WERs, averaged over the number of utterances in the test set.} using text+speech input. Our ASR prediction system is built using both \textit{Keras} \cite{chollet2015keras} and \textit{Tensorflow\footnote{https://www.tensorflow.org}}. 

In the next section, we analyze the representations learnt in the higher layers (3 blocks colored in yellow and dotted in Figure~\ref{fig:architecture}) for pure text (TXT), pure speech (RAW-SIG) and both (TXT+RAW-SIG).
 
\section{\label{sec:EvaluatingLR}Evaluating learned representations}
\subsection{Methodology}
In this section, we attempt to understand what our best ASR performance prediction system \cite{elloumi2018asr} learned.
We analyze the text and speech representations obtained by our architecture. 
Alike  \cite{belinkov2017analyzing}, the joint text+speech model is used to generate utterance level features (hidden representations of speech turns colored in yellow in Figure 
 \ref{fig:architecture}) that are given to a shallow classifier trained to solve secondary classification tasks such as:
 
\begin{itemize}

\item  \textbf{STYLE:} classify the utterances between (\textit{spontaneous} and \textit{non spontaneous}) styles (see table \ref{tab:data}), 
\item  \textbf{ACCENT:} classify the utterances between \textit{native} and \textit{non native} speech (see also table \ref{tab:data}, we used the speaker annotations provided with our datasets in order to label our utterances in native/non native speech),
\item  \textbf{SHOW:} classify the utterances in different broadcast programs (as described in table \ref{tab:DistShow}, each utterance of our corpus is labeled with a \textit{broadcast program name}).

\end{itemize}

As a more visual analysis, we also plot an example of hidden representations projected to a 2-D space using t-distributed Stochastic Neighbor Embedding (t-SNE) \cite{maaten2008visualizing}.\footnote{\url{https://lvdmaaten.github.io/tsne/code/tsne_python.zip}}

\subsection{\label{subsec:classifier}Shallow classifiers}
We built three shallow classifiers (SHOW, STYLE, ACCENT) with a similar architecture. The classifier is a feed-forward neural network with one hidden layer (size of the hidden layer is set to 128) followed by dropout (rate of 0.5) and a ReLU non-linearity. Finally,  a $softmax$ layer is used for mapping onto the label set size. We chose this simple formulation as we are interested in evaluating the quality of the representations learned by our ASR prediction model, rather than optimizing the secondary classification tasks. The network input size depends on which layer to analyze (see figure~\ref{fig:architecture}).  
Training is performed using $Adam$ \cite{DBLP:journals/corr/KingmaB14} (using default parameters) over shuffled mini-batches in order to minimize the cross-entropy loss. The models are trained for 30 epochs with a batch size of 16 speech utterances. After training, we keep the model with the best performance on DEV set and report its performance on the TEST set. The classifier outputs are evaluated in terms of accuracy.

\subsection{\label{data}Data} 

A data set from \cite{elloumi2018asr} was employed in our
experiments, divided into three subsets: training (TRAIN), development (DEV) and test (TEST). Speech utterances come from various French broadcast collections gathered during projects or shared tasks: \textit{Quaero},  \textit{ETAPE}, \textit{ESTER 1 \& ESTER 2} and  \textit{REPERE}.
 
The TEST set  contains unseen broadcast programs that are different from those present in TRAIN and DEV  \cite{elloumi2018asr}.

 \begin{table}[thb] 
  \centering
 
   \small
  \begin{tabular}{lccc}
    \toprule
       \textbf{Category}  & \textbf{TRAIN} & \textbf{DEV}   & \textbf{TEST} \\
    \midrule 
    Non Spontaneous     &54250 & 6101 & \textbf{3109}\\
    Spontaneous    &\textbf{13277} & \textbf{1403} & 3728 \\ 
    \hline
    Native     & 44487 &  4945 & 5298 \\
    Non Native    & \textbf{23040} &  \textbf{2559} & \textbf{1539} \\ 
    \bottomrule
  \end{tabular}
   \captionof{table}{Distribution of our utterances between non spontaneous and spontaneous styles,  native and non native accents}
       \label{tab:data}
 \end{table}
 
\begin{table}[!ht]
  \centering
  \small
 
  \begin{tabular}{lcc}
    \toprule  
    \textbf{Show} & \textbf{TRAIN} &  \textbf{DEV} 
  \\   \midrule 
 FINTER-DEBATE &7632 & 833 
 \\  FRANCE3-DEBATE&928 & 77   
 \\  LCP-PileEtFace&\textbf{4487}& 525
 \\  RFI & 25565 & 2831
 \\  RTM & 24198 &2745
 \\  TELSONNE& 4717 & \textbf{493}
 \\  \hline  \textbf{Total}& 67527 & 7504
 \\ \bottomrule
\end{tabular}
 \captionof{table} {Number of  utterances for each broadcast program}
       \label{tab:DistShow}
 \end{table}
 Tables \ref{tab:data} and \ref{tab:DistShow} show the whole data set in terms of speech turns available for each classification task. We clearly see that the data is unbalanced for the three categories (STYLE, ACCENT, SHOW).
Since we are interested in evaluating the discriminative power of our learned representations for these 3 tasks, we extracted a balanced version of our TRAIN/DEV/TEST sets by filtering among over-represented labels (final number of kept utterances corresponds to bold numbers in table \ref{tab:data} and \ref{tab:DistShow} ).  
 Table~\ref{tab:balanced_data} shows  the distribution of our final balanced TRAIN/DEV/TEST sets as well as the number of categories for each task.\footnote{For the \textit{SHOW} classification task, the \textit{FRANCE3-DEBATE} shows were finally removed since they represent a too small amount of speech turns.}

 \begin{table}[!ht]
 \centering

\small
\begin{tabular}{ lcccc  }
  \hline
  \multirow{2}{*}{} &  \multirow{2}{*}{\textbf{\#Catg}}&  \multicolumn{3}{c}{\textbf{Turns of speech per category}}   \\
  & &  \textbf{  TRAIN}  &  \textbf{ DEV} &  \textbf{TEST}
  \\\hline \textbf{SHOW}  & 5 & 4487$_{\times5}$   & 493$_{\times5}$ &  -
  \\ \textbf{STYLE} &2 & 13277$_{\times2}$   & 1403$_{\times2}$ &  3109$_{\times2}$ 
  \\ \textbf{ACCENT} & 2&   23040$_{\times2}$   & 2559$_{\times2}$ &  1539$_{\times2}$ 
 \\\bottomrule 
\end{tabular}
 \captionof{table}{Description of our balanced data set for each category}
\label{tab:balanced_data}
 \vspace{-.5cm}
\end{table} 

\subsection{Results}
For each classification task, we build a shallow classifier using the hidden representations of \textit{TXT}, \textit{RAW-SIG} and \textit{TXT+RAW-SIG} blocks as input.  The experimental results are presented in table \ref{tab:Perf-Classification} for both DEV and TEST sets separated by two vertical bars ($||$). 

Classification performance is all above a random baseline accuracy ($>$50\% for STYLE and ACCENT and $>$20\% for SHOW). This shows that training a deep WER prediction system gives representation layers that contain a meaningful amount of information about speech style, speech accent and broadcast program label. Predicting utterance style (spontaneous/non spontaneous) is slightly easier than predicting accent (native/non native) especially from text input. One explanation might be that   speech utterances are short ($<6s$) while accent identification needs probably longer sequences. We also observe that using both text and speech improves the learned representations for the STYLE task while it is less clear for the ACCENT task (for which improvement seen on DEV is not confirmed on TEST).
\zied{Finally, text input is significantly better than speech input whereas we could have expected better performance from speech for the SHOW task (speech signals convey information about the audio characteristics of a broadcast program). 
It means that text input contains  correlated  information with broadcast-program type, speech style and speaker's accent.
In case of SHOW task, our performance prediction system is able to capture information (vocabulary, topic, syntax, etc.) about a specific broadcast program type, based on textual features and to differ it from others (radio programs, TV debate programs, phone calls, broadcast news programs, etc.). Likewise,  the textual information captured is very different between spontaneous/non-spontaneous speech styles and native/non-native speaker's accents. }

Among the representations analyzed, the outputs of the CNNs (A1,B1) lead to the best classification results, in line with previous findings about convolutions as feature extractors. Performance then  drops using the higher (fully connected) layers that do not generate better representations for detecting style, accent or show.

 \begin{table}[!ht]
\footnotesize
 \centering

\begin{tabular}{lcccc}
   
    \toprule
  \textbf{Layer} & \textbf{Dim.} &\textbf{SHOW} & \textbf{STYLE} & \textbf{ACCENT} \\  \midrule 
  \multicolumn{5}{c}{ \cellcolor{lightgray!50}  TXT }   \\ 
   A1 & 1280 & \textbf{57.12}$||$-&\textbf{80.72}$||$68.99&\textbf{70.75}$||$66.54  \\   
   A2  & 256 & 54.89$||$-&80.01$||$\textbf{69.56 }&69.30$||$69.43  \\  
   A3   &  128 & 51.04$||$-& 79.23$||$68.27 &68.25$||$\textbf{70.89}\\     
   \multicolumn{5}{c}{ \cellcolor{lightgray!50} RAW-SIG}  \\    
   B1  & 512 & \textbf{42.35}$||$-&  \textbf{72.92}$||$\textbf{58.64}&64.60$||$\textbf{55.85}\\    
   B2   &  512 & 41.22$||$-& 72.20$||$58.41&64.44$||$54.84\\    
   B3  &  256  & 41.22$||$- & 72.38$||$58.44 &64.50$||$54.65\\  
   B4    &  128 & 40.77$||$- & 72.38$||$58.52&\textbf{64.74}$||$54.87\\  
   
   \multicolumn{5}{c}{\cellcolor{lightgray!50} TXT + RAW-SIG }   \\   
   C1 {\tiny \textbf{(A3+B4)}} &  256  &\textbf{57.04}$||$-& \textbf{81.29}$||$70.36&\textbf{71.41}$||$\textbf{65.98}\\    
   C2 &  128  &53.06$||$-& 79.62$||$\textbf{70.55}&70.01$||$65.20\\  \hline
   \hline \textbf{Random} & - &\textbf{20.00} & \textbf{50.00}  &\textbf{50.00}
\\\bottomrule 
\end{tabular}
 \captionof{table}{Show/Style/Accent classification accuracies using representations from different layers learned during the training of our ASR WER prediction system.}
 \label{tab:Perf-Classification}
 \vspace{-.3cm}
\end{table}

    \begin{figure}[ht!]
   \begin{minipage}[b]{0.48\linewidth}   
      \centering \includegraphics[scale=0.25]{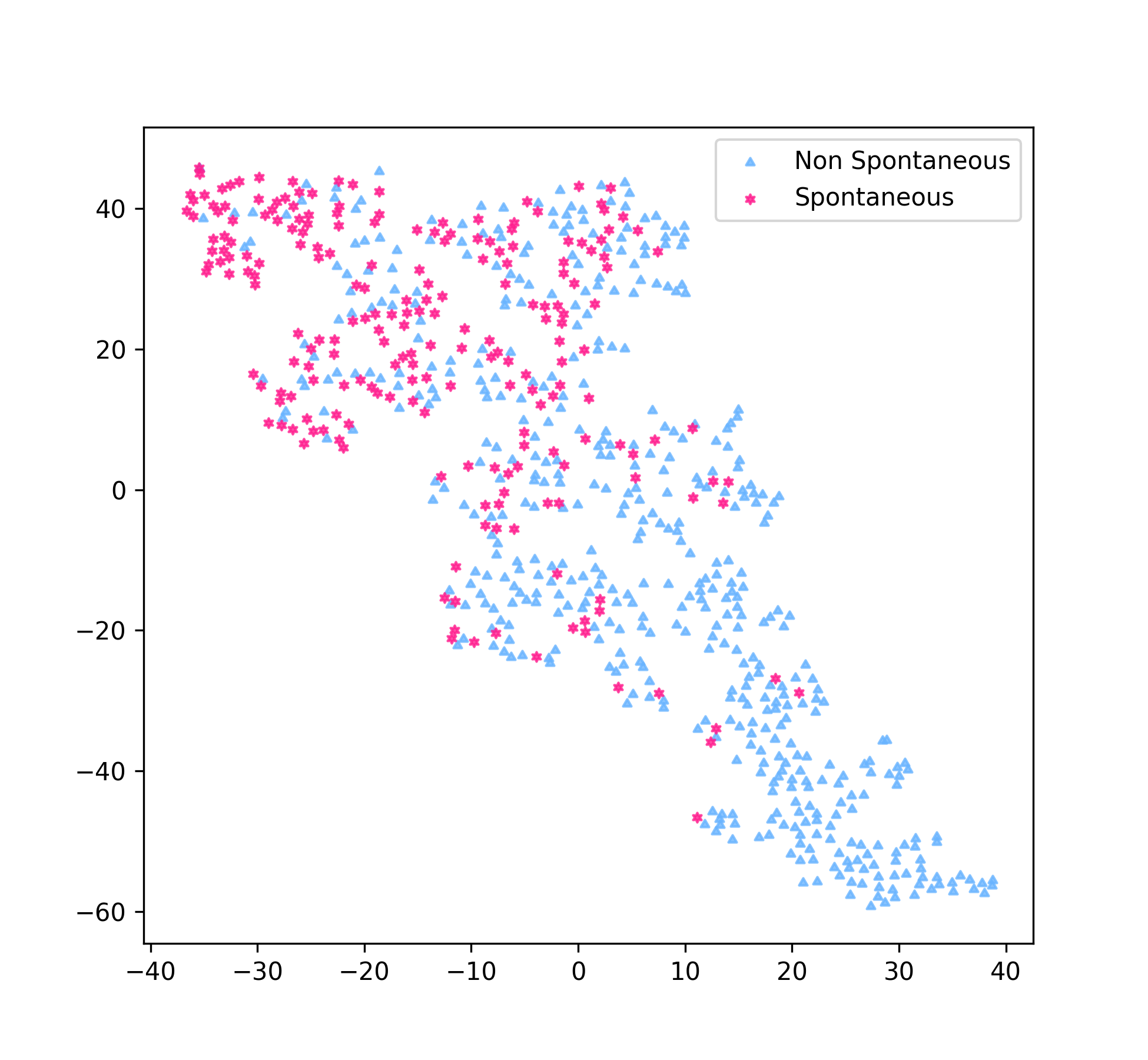}
      \caption*{ (a) 4s$\leq$D$<$5s   }
   \end{minipage}
      \begin{minipage}[b]{0.48\linewidth}
      \centering \includegraphics[scale=0.25]{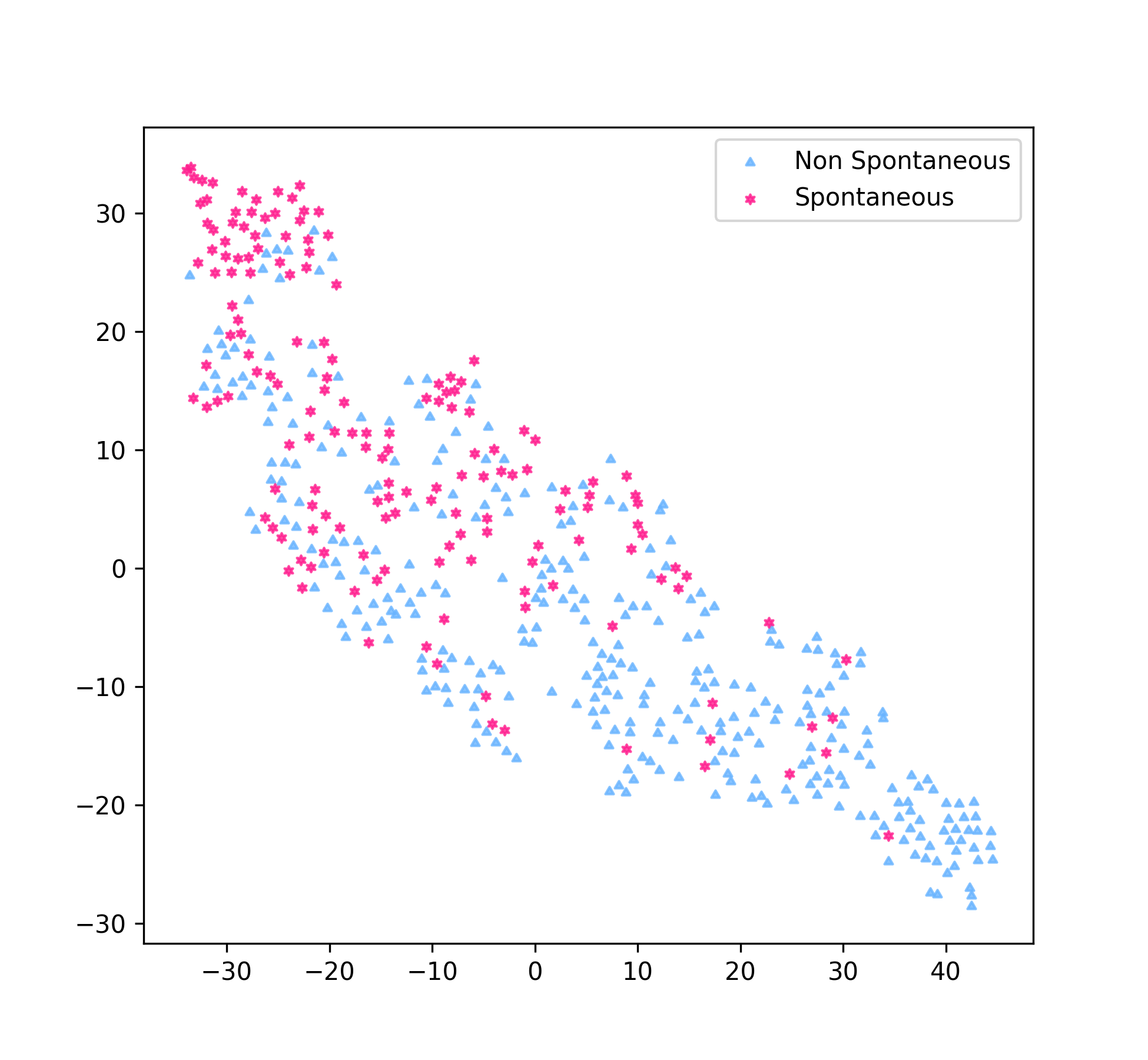}
      \caption*{ (b) 5s$\leq$D$<$6s   }
   \end{minipage}\hfill
    \caption{Visualization of utterance representations from C2 layer for different speech styles (S spontaneous - NS non spontaneous) - (a) utt. length is 4s$\leq$D$<$5s  and (b)  5s$\leq$D$<$6s}
    \label{fig:tsne}
    \vspace{-.3cm}
   \end{figure}
 
We visualize an example of utterance representations from C2(TXT+RAW-SIG) layer in figure \ref{fig:tsne} using the t-SNE. For a fixed utterance duration   4s$\leq$D$<$5s (716 speech turns) and 5s$\leq$D$<$6s   (489 speech turns), non spontaneous utterances are plotted in blue while spontaneous ones are in pink.
The C2 layer produces clusters which shows that spontaneous utterances are in the upper-left part of the 2D space. This suggests that C2 hidden representation captures a weak signal about speaking style.

\begin{figure}[!ht] 
      \centering \includegraphics[width=\linewidth]{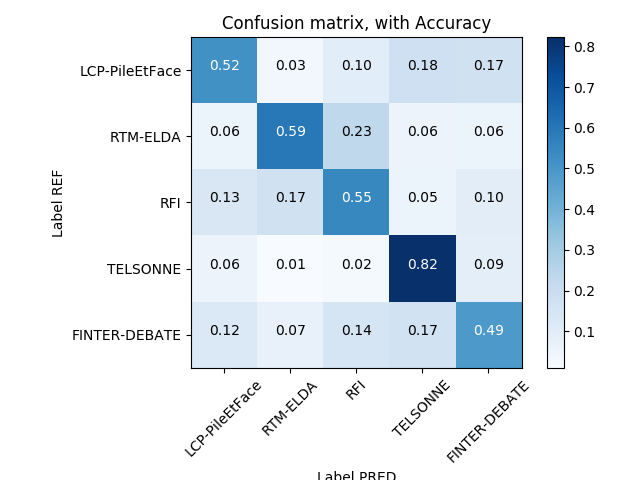}
      \caption{Confusion matrix for SHOW classification using C2(TXT+RAW-SIG) layer as input, evaluated on DEV}
   \label{fig:matrix}
    \vspace{-.3cm}
   \end{figure}

Finally, figure ~\ref{fig:matrix}  is the confusion matrix  produced using C2(TXT+RAW-SIG) layer. The classifiers  very well predicted \textit{TELSONNE} category (Accuracy of 82\%), which contains many phone calls from the radio listeners. This show is rather different from the 4 other shows in DEV (broadcast debates and news).

  \begin{table*}[!ht]
\small
 \centering

\begin{tabular}{lccccc }
  \toprule

 \multirow{2}{*}{\textbf{Models}} & \multicolumn{2}{c }{ \textbf{Performance prediction task }} & \multicolumn{3}{c }{ \textbf{Classification tasks }}
 \\ &  \textbf{MAE}  & \textbf{Kendall} 
 &   \textbf{SHOW}  & \textbf{STYLE} & \textbf{ACCENT}

 \\\midrule 
 
\rowcolor{lightgray!50} &   \multicolumn{2}{c}{\textbf{Baseline: Mono-task} } & & &
\\  \textbf{WER} \cite{elloumi2018asr} & 15.24$||$19.24 & 45.00$||$46.83   &- &- &-

 \\ \rowcolor{lightgray!50} & \multicolumn{2}{c} {\textbf{2-task}}   &  & &
\\\textbf{WER SHOW}& \textbf{14.83}$||$19.15  &    \textbf{47.25}$||$47.05  &\textbf{99.29}$||$-  & - & - 
\\\textbf{WER STYLE} &15.07$||$19.66 & 45.92$||$45.49 &-&  99.01$||$65.24 &-
\\\textbf{WER ACCENT}& 15.05$||$19.60 & 46.17$||$45.60&-&-& 91.72$||$ 75.30 

 \\\rowcolor{lightgray!50}  &\multicolumn{2}{c}{ \textbf{3-task}} &  & &  
 \\\textbf{WER STYLE ACCENT} & 15.12$||$20.23 & 45.75$||$44.09 & - & 98.63$||$69.07 & 88.99$||$ \textbf{77.46}
\\\textbf{WER SHOW ACCENT} & 14.94$||$19.76 & 46.19$||$43.61 & 98.38$||$- & - & 89.87$||$71.44
\\\textbf{WER SHOW STYLE} & 14.90$||$\textbf{19.14} & 45.87$||$\textbf{47.28} &99.12$||$-   & \textbf{99.47}$||$\textbf{81.98} & -

 \\ \rowcolor{lightgray!50} &\multicolumn{2}{c}{\textbf{4-task}}  && &
\\\textbf{WER SHOW STYLE ACCENT}& 15.15$||$19.64 & 45.59$||$45.42 & 99.04$||$- &99.29$||$81.55 & \textbf{91.92}$||$73.60
\\\hline \hline 
\textbf{WER ALL COMBINED OUTPUTS}& \textbf{14.50} $||$\textbf{18.87} &\textbf{ 48.16}$||$\textbf{48.63} & - &- &-

\\\bottomrule 
\end{tabular}
 \captionof{table}{Evaluation of ASR performance prediction with multi-tasks models ($DEV||TEST$) computed with MAE and Kendall - secondary classification tasks accuracy is also reported}

\label{tab:PerfMultiTask}
 \vspace{-.3cm}
\end{table*}
\section{\label{sec:multi-task}Multi-task learning}

We have seen in the previous section that, while training an ASR performance prediction system, hidden layers convey a clear signal about speech style, accent and show. This suggests that these 3 types of information might be useful to structure the deep ASR performance prediction models. 
\zied{In this section, we investigate the effect of knowledge of these labels (style, accent, show) at training time on prediction systems qualities.} 
For this, we perform multi-task learning providing the additional information about broadcast type, speech style and speaker's accent during training.
The architecture of the multi-task model is similar to the  single-task WER prediction model of Figure \ref{fig:architecture} but we add additional outputs: a $softmax$ function is added for each new classification task after the last fully connected layer (C2). The output dimension depends on the task: 6 for SHOW  and 2  for STYLE and ACCENT tasks.

We use the full (unbalanced) data set described in tables \ref{tab:data} and \ref{tab:DistShow}. Training of the multitask model uses $Adadelta$ update rule and all parameters are initialized from scratch (8.70M). Models are performed for 50 epochs with batch size of 32. MAE  is used as the loss function for WER prediction task while cross-entropy loss is used for the classification tasks.

In the composite (multitask) loss, we assign a weight of 1 for MAE loss (main task) and a smaller weight of 0.3 \zied{(tuned using a grid search on DEV dataset)} for cross-entropy (secondary classification task) loss(es). 

After training, we take the model that lead to the best \textit{MAE} on \textit{DEV} set and report its performance on TEST. 
We build several models that simultaneously address 1, 2, 3 and 4 tasks. The models are evaluated with a specific metric for each task: MAE \& Kendall\footnote{Correlation between true ASR values and predicted ASR values} for WER prediction task  and Accuracy for classification tasks.

Table \ref{tab:PerfMultiTask} summarizes the experimental results on DEV and TEST sets, separated by two vertical bars ($||$). We considered the mono-task model described in  \cite{elloumi2018asr} (and summarized in section 3) as a baseline system. 

We recall that we evaluated the SHOW classification task only on the DEV set (TEST broadcast programs are new and were unseen in the TRAIN).
 
First of all, we notice that performance of classification tasks in muti-task scenarios are very good: we are able to train efficient ASR performance prediction systems that simultaneously tag the analyzed utterances according to their speech style, accent and broadcast program origin. Such multi-task systems might be useful diagnostic tools to analyze and predict ASR on large speech collections. 
Moreover, our best multi-task systems display a better performance (MAE, Kendall) than the baseline system, which means that the implicit information given about style, accent and broadcast program type can be helpful to structure the system's predictions.
For example, in 2-task case, the best model is obtained on WER+SHOW tasks with a difference of +0.41\%, +2.25\% for MAE and Kendall respectively (on DEV) compared to the baseline on WER prediction task. 
However, it is also important to mention that the impact of multi-task learning on the main task (ASR performance prediction) is limited: only slight improvements on the test set are observed for MAE and Kendall metrics. Anyway, the systems trained seem complementary since their combination (averaging, over all multi-task systems, predicted WERs at utterance level) leads to significant performance improvement (MAE and Kendall).

\section{\label{sec:conclusion}Conclusion}  
This paper presented an analysis of  learned representations of our deep ASR performance prediction system. Experiments show that hidden layers convey a clear signal about speech style, accent, and broadcast type.  
We also proposed a multi-task learning approach to simultaneously predict WER and classify utterances according to style, accent and broadcast program origin.
\bibliographystyle{acl_natbib_nourl}
\bibliography{emnlp2018}
\end{document}